\documentclass[10pt,twocolumn,letterpaper]{article}

\usepackage{cvpr}              % To produce the CAMERA-READY version
\pdfoutput=1
\usepackage{amsmath,amssymb} % define this before the line numbering.
\usepackage{multirow}
\usepackage{gensymb}
\usepackage{makecell}
\usepackage{float}
\usepackage[dvipsnames]{xcolor}

\DeclareUnicodeCharacter{0301}{\'{e}}
\definecolor{cvprblue}{rgb}{0.21,0.49,0.74}
\usepackage[pagebackref,breaklinks,colorlinks,citecolor=cvprblue]{hyperref}

\title{Geometry-Aware Feature Matching for Large-Scale Structure from Motion}

\author{
    Gonglin Chen$^{1, 2}$ \and
    Jinsen Wu$^{1, 2}$ \and
    Haiwei Chen$^{1, 2}$ \and
    Wenbin Teng$^{1, 2}$ \and
    Zhiyuan Gao$^{1, 2}$ \and
    Andrew Feng$^{1}$ \and
    Rongjun Qin$^{3}$ \and
    Yajie Zhao$^{1, 2}$\thanks{Corresponding author.}\\
    \and
    {\tt\small \{gonglinc, jinsenwu\}@usc.edu, chw9308@hotmail.com,}
    {\tt\small \{wenbinte, gaozhiyu\}@usc.edu}\\
    {\tt\small andrewfengusa@gmail.com, qin.324@osu.edu, zhao@ict.usc.edu}
    \\
    \and
    $^{1}$Institute for Creative Technologies
    \and
    $^{2}$University of Southern California 
    \and
    $^{3}$The Ohio State University 
}

\begin{document}
\maketitle
% \maketitlesupplementary
\begin{abstract}
%Establishing consistent and dense correspondences across multiple images is crucial for Structure from Motion (SfM) systems. Significant view changes, which are common in in-the-wild images, pose an even greater challenge to the correspondence solvers. We present a novel optimization-based approach that integrates detector-based and detector-free methods to address these challenges effectively. Our technique starts with utilizing sparse correspondences from detector-based methods as anchor points. Utilizing these anchor points as priors, the optimization steps then perform correspondence refinement to the dense features obtained from detector-free methods, under geometric constraints formulated by the Sampson Distance. This hybrid approach increases the density and accuracy of the correspondences in large-scale outdoor scenes and mitigates multi-view inconsistencies commonly found in detector-free approaches. Extensive experimental evaluations have demonstrated that our method achieves significant improvements on the camera pose accuracy and the reconstructed point cloud density compared to previous state-of-the-art methods.

Establishing consistent and dense correspondences across multiple images is crucial for Structure from Motion (SfM) systems. Significant view changes, such as air-to-ground with very sparse view overlap, pose an even greater challenge to the correspondence solvers. We present a novel optimization-based approach that significantly enhances existing feature matching methods by introducing geometry cues in addition to color cues. This helps fill gaps when there is less overlap in large-scale scenarios. Our method formulates geometric verification as an optimization problem, guiding feature matching within detector-free methods and using sparse correspondences from detector-based methods as anchor points. By enforcing geometric constraints via the Sampson Distance, our approach ensures that the denser correspondences from detector-free methods are geometrically consistent and more accurate. This hybrid strategy significantly improves correspondence density and accuracy, mitigates multi-view inconsistencies, and leads to notable advancements in camera pose accuracy and point cloud density. It outperforms state-of-the-art feature matching methods on benchmark datasets and enables feature matching in challenging extreme large-scale settings. Project page: \href{https://xtcpete.github.io/geo-website/}{https://xtcpete.github.io/geo-website/}.

\begin{figure}[!ht]
    \centering
    \includegraphics[width=1\linewidth]{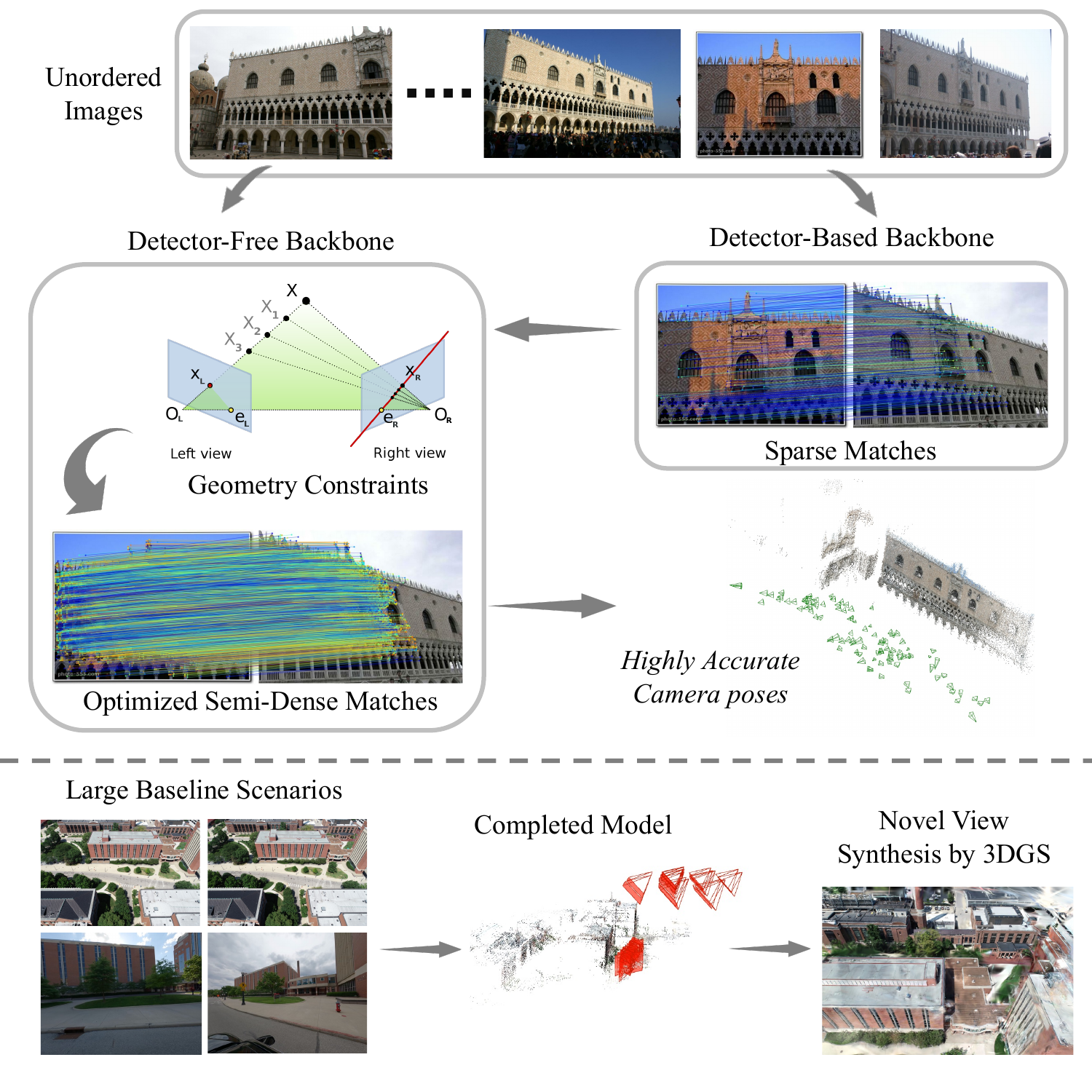}
    \caption{Our proposed method bridges detector-based feature matching with detector-free feature matching. We utilize sparse correspondences as geometric priors to iteratively optimize the matching process. By leveraging the denser matches from detector-free matchers, our method achieves highly accurate camera pose recovery and generates denser point clouds. This approach is particularly effective in challenging large baseline scenarios, such as air-to-ground imagery, and provides substantial benefits to downstream applications like novel view synthesis.}
    \label{fig:teaser}
    \vspace{-1em}
\end{figure}
\end{abstract}
    
\vspace{-1em}
\section{Introduction}

% This problem we study is important
Correspondence, as perhaps the most fundamental problem in computer vision, helps us with mapping the world. Solving correspondence is required in most applications in Augmented Reality and Robotics since it derives the camera poses and 3D point locations~\cite{kerbl3Dgaussians}.

% The problem is hard in the application setting we want (aka 3D recon)
In the 3D reconstruction from a set of uncalibrated images, the correspondence problem meets two additional demands: 1) the demand to be able to observe a corresponding point across multiple views, which is known as the \textit{track length} of a correspondence point, and 2) the demand to find a \textit{dense} set of such points such that they are sufficient for the recovery of 3D geometry of the scene. 

\begin{figure}[!ht]
    \centering
    \includegraphics[width=0.98\linewidth]{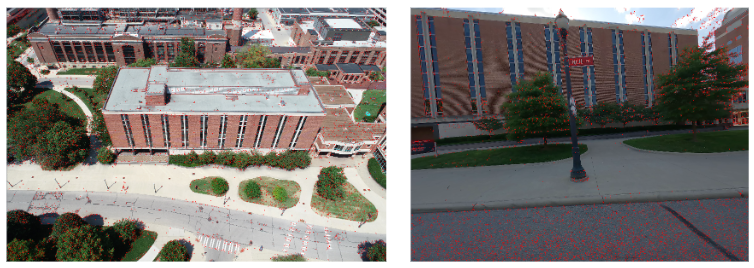}
    \caption{Challenging case of detector-based and detector-free feature matching methods for cross view images.}
    \label{fig:detector}
    \vspace{-0.5 em}
\end{figure}

In this paper, we consider a very difficult correspondence problem—dense, consistent correspondences across a sequence of frames with large camera baselines. To our knowledge, no prior works have achieved satisfactory results in the presence of these challenges: both detector-based and detector-free methods struggle with significant view changes, as illustrated in Figure~\ref{fig:detector}, due to their inherent limitations. Detector-based methods often match only a small number of keypoints, making accurate image registration difficult. 

While the 3D reconstruction problem described thereof has been well studied for decades, they are notably tackled by either requiring a pre-calibration of the captured cameras~\cite{5940405,1640781}, or the requirement that consecutive frames captured are closely located to each other~\cite{4160954,8593691,6126513,6619022}. In recent years, however, correspondence methods based on learned artificial neural networks (ANNs) have demonstrated significant improvements over the accuracy and robustness of the matching between a pair of in-the-wild, uncalibrated images~\cite{superpoint2018,sarlin20superglue,zhao2023aliked,lindenberger2023lightglue,edstedt2023dkm,sun2021loftr,chen2022aspanformer}, which has given rise to the possibility of reconstructing 3D scenes from the abundant, in-the-wild photos captured in daily setting~\cite{5459148, schoenberger2016sfm, mohr1995relative, polic2018fast}.

Recent state-of-the-art correspondence methods are separated into two major branches: the detector-based methods~\cite{Lowe:2004:DIF:993451.996342,superpoint2018,sarlin20superglue} that employ two independent steps to extract feature descriptor for key points first and perform matching later, and the detector-free methods~\cite{sun2021loftr,chen2022aspanformer,edstedt2023dkm,wang2022matchformer} that find matches in an end-to-end trained neural networks in one shot. While the detector-based methods excel in performing matching among image sets with rich texture and small view point changes, their reliance on the keypoint detection and description poses significant challenges when images of interest lack textures, or their local features are not sufficient for matching.

\begin{figure}[!ht]
    \centering
    \includegraphics[width=1\linewidth]{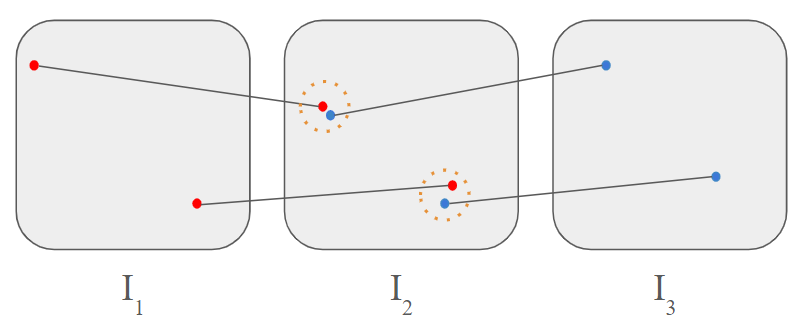}
    \caption{Matches from detector-free methods suffered from the problem of short track length (the track breaks at $I_2$). The location of the matches in $I_2$ depends on the images that it pairs with. }
    \label{fig:track_length}
    \vspace{-0.5 em}
\end{figure}

On the other hand, the detector-free methods~\cite{sun2021loftr,chen2022aspanformer,wang2022matchformer,edstedt2023dkm} naturally find much denser correspondences between a pair of images with a dense search, but they typically lack control of the locations of keypoints across views and are optimized for sub-pixel correspondences that depend on specific image pairs. As shown in Figure~\ref{fig:track_length}, this pair dependency makes them unsuitable for maintaining long-track correspondence across multiple images. Therefore, it leads to poor performance when these methods are directly applied to Structure-from-Motion (SfM) reconstruction. Though quantization on the found matches can mitigate the problem of short track length, it comes at the price of accuracy and requires additional refinement that makes the entire system more computationally expensive. 

We aim to bridge the gap between the detector-based and detector-free methods in order to obtain \textbf{dense} correspondences that can be tracked \textbf{across multiple views} with \textbf{large baselines} - addressing a very difficult correspondence problem in SfM reconstruction from in-the-wild image collections. Specifically, we propose a novel optimization-based method that refines the pair-wise dense correspondences by iteratively filtering and reassigning correspondence under a geometric constraint formulated by the Sampson Distance~\cite{Hartley:2003:MVG:861369}. This method allows the early rejection of outliers during the matching steps, improving the reliability of correspondences in large-baseline scenarios. 
Interestingly, we have observed that incorporating both priors from the existing detector-based method and enforcing the geometric constraint helps our model achieve better camera prediction and longer track length among the found correspondences, as demonstrated in Section~\ref{ablation}. 
The designed optimization module thus consistently improves upon any detector-based and detector-free backbones, by providing denser correspondences compared to the detector-based methods, while refining the detector-free results to be more accurate and consistent across multiple views. 3D point cloud reconstructed with our feature matcher thus excels in both density and accuracy compared to previous state-of-the-art approaches.

To validate the effectiveness of our proposed method, we have conducted comprehensive evaluations using several publicly available datasets, including Image Matching Competition Benchmark~\cite{Jin2020} and ScanNet~\cite{dai2017scannet}. We compared the SfM reconstruction results using matches 
%\haiwei{reconstruction results? or correspondence results?} 
from our feature matcher against several state-of-the-art baseline methods. The 
%\Jinsen{"our" instead of "the"?} 
results demonstrated better accuracy in estimated camera poses and longer track length compared to those from the baselines.
%\haiwei{only camera pose accurarcy?} .
Additionally, we have extended our evaluation to a challenging large-baseline setting that requires mapping from the UAV to street view perspectives (Figure ~\ref{fig:air_ground}).
\section{Related work}
\label{sec:relatedWork}

\begin{figure*}
  \centering
   \includegraphics[width=1.0\linewidth]{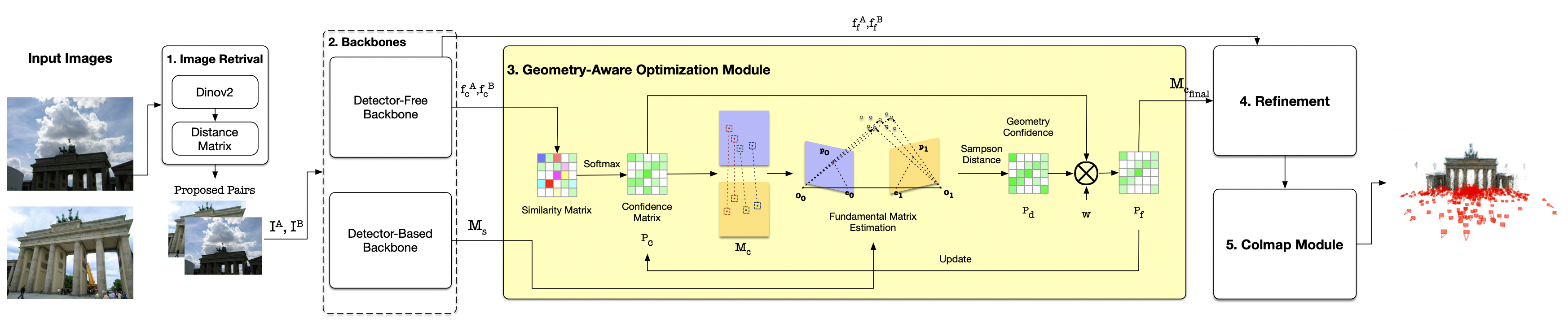}
   \vspace{-1 em}
   \caption{An overview of our pipeline for SfM reconstruction. \textit{\textbf{1.}} the pipeline runs image retrieval based on global embeddings generated by dinov2~\cite{oquab2024dinov2}. \textit{\textbf{2}}. A backbone module takes image pairs as input.  The image pairs will be processed by a Detector-Free Backbone and a Detector-Based backbone. \textit{\textbf{3.}} A geometry-aware optimization module is applied to iteratively optimize the fundamental matrix and matches with anchor points from detector-based methods. \textit{\textbf{4.}} The final matched coarse points are refined using a correlation-based refinement block. \textit{\textbf{5.}} Final refined matches are then fed into COLMAP~\cite{schoenberger2016sfm,schoenberger2016mvs} for SfM.}
   \label{fig:pipeline}
    \vspace{-1em}
\end{figure*}

\subsection{Detector-based Feature Matching} A conventional detector-based matching framework typically detects keypoints in images, describes them using high-dimensional vectors, and further applies a matching algorithm to find the correspondences between these image sets~\cite{KeypointRGBD2016}. Historically, numerous handcrafted methods like SIFT~\cite{Lowe:2004:DIF:993451.996342}, SURF~\cite{SURF2008}, and ORB~\cite{ORB2011} have been used for the keypoint detection.  Lately, learning-based methods~\cite{superpoint2018, sarlin20superglue, LIFT2016}, particularly those leveraging convolution neural networks (CNNs), have significantly improved the performance over the handcrafted features and shown robustness under large viewpoint and illumination changes. For instance, SuperPoint~\cite{superpoint2018} achieved a good performance by creating a vast dataset of pseudo-ground-truth interest points and using a self-supervised interest point detector. SuperGlue~\cite{sarlin20superglue} learns feature matching using point features from a graph neural network (GNN) and attention layers that integrate global context information. 

Patch2Pix~\cite{ZhouCVPRpatch2pix} has made further improvement with patch-level matches and geometry-guided refinement that has enhanced the correspondence accuracy. Patch2Pix~\cite{ZhouCVPRpatch2pix} utilizes the Sampson distance~\cite{Hartley:2003:MVG:861369} to refine initial patch-level matches to pixel-level accuracy, ensuring geometric consistency and precision. However, Patch2Pix~\cite{ZhouCVPRpatch2pix} is limited to refining existing sparse matches and add new matches or extend beyond the initial patch-level matches.
Despite their success, detector-based methods are limited by their dependency on feature detectors, which struggle to consistently identify distinct interest points in areas that lack clear and distinguishable features.  

\subsection{Detector-free Feature Matching} Detector-free approaches~\cite{Huang2023adamatcher,sun2021loftr,chen2022aspanformer,schmidt2016RAL,choy2016unc} bypass the feature detection stage and directly generate matches, following a concept that originates from SIFT Flow~\cite{siftflow}.  The introduction of learning-based methods that employ contrastive loss to train pixel-wise feature descriptors~\cite{schmidt2016RAL, choy2016unc} marked a significant development in this field. These methods - similar to detector-based approaches - employ the nearest neighbor search for post-processing to match the dense descriptors. 

NCNet~\cite{NCNet2018} learns dense correspondences in an end-to-end manner, creating 4D cost volumes that account for all possible inter-image matches. 4D convolutions are applied to refine these volumes, promoting consistency across the correspondence field.  Building on this work, SparseNCNet~\cite{sparseNCNet2020} improved the efficiency with sparse convolutions. DRCNet~\cite{DRCNet2020} improved dense matching accuracy with a coarse-to-fine architecture.  While NCNet addresses all potential matches, its 4D convolution’s receptive field is limited to small local neighborhoods.  

LoFTR~\cite{sun2021loftr}, in contrast, aims for a broader match consensus, leveraging the expansive receptive field of transformers.  ASpanFormer~\cite{chen2022aspanformer}, MatchFormer~\cite{wang2022matchformer}, and Efficient LoFTR~\cite{wang2024eloftr} follows the transformer-based design to utilize multilevel cross-attention to mutually update image features, thereby incorporating dual-view contexts for achieving both global and local consensus.  While these methods represent significant improvements to the state-of-the-art, they still struggle to uphold high-level contexts such as epipolar geometries.  Therefore, we propose a geometry-aware approach that enforces epipolar consistency in dense feature matching.

 \subsection{Structure from Motion} Structure from Motion (SfM) methods have been extensively studied~\cite{mohr1995relative,10.1007/3-540-61123-1_181} in the past. Some of these efforts focus on the scaling to large-scene reconstruction~\cite{schoenberger2016sfm, 5459148, schoenberger2016mvs}. SfM methods mainly use detector-based matching at the beginning of their pipelines, which limits the overall SfM performance in challenging cases. Some end-to-end SfM methods propose to directly model camera poses~\cite{vijayanarasimhan2017sfmnet,parameshwara2022diffposenet,zhang2022relpose}. These methods avoid the bottleneck of feature detection but have limited scalability to real-world scenarios. Although Recent differentiable rendering methods~\cite{mildenhall2020nerf,kerbl20233d} can optimize poses and the scene representation simultaneously, they rely on the traditional SfM pipelines like COLMAP\cite{schoenberger2016sfm,schoenberger2016mvs} to provide the initial camera poses. Lindenberger et al. ~\cite{lindenberger2021pixsfm} improves the accuracy of SfM and visual localization by refining keypoints, camera poses, and 3D points post to the traditional SfM pipelines. Unlike previous SfM methods, our proposed feature matcher ensures robustness and local geometry consistency at the first step of an SfM system.
\section{Methods}
\label{methods}

The core of our methods is an optimization module that operates on existing sparse matches $M_s$ and two feature maps $f_c^{A}$ and $f_c^{B}$ from a pair of images. The module optimizes for the relative camera poses between the two images and refines the input coarse correspondences into refined coarse correspondence $M_{c_{final}}$ that is geometric consistent under a Sampson distance~\cite{Hartley:2003:MVG:861369} constraint. In Section~\ref{sec:prelim}, we provide an overview of how the sparse matches and feature maps are obtained, as well as the definition of Sampson distance. In Section~\ref{feature matching}, we discuss the details of our optimization module.

\subsection{Preliminaries}
\label{sec:prelim}

\subsubsection{Detector Based Method} 
Given two images $I^A$ and $I^B$, our method accepts a detector-based backbone $F_{db}$ that first identifies the sparse interest points with a keypoint detector. Then, local features are extracted around the identified interest points, encoding the local visual appearance in high-dimensional vectors. These features are then matched between $I^A$ and $I^B$ using either learned neural networks~\cite{sarlin20superglue, lindenberger2023lightglue} or nearest-neighbor search to generate a set of sparse correspondences $M_s$. In our implementation, we use SuperPoint \cite{superpoint2018} + SuperGlue \cite{sarlin20superglue} to obtain sparse correspondence between a given image pair as it has been shown that the two methods combined can provide anchor points that introduce robust and good priors to guide dense matching~\cite{Kuang2022DenseGAP}. We also evaluate the performance of different detector-based backbones in Section~\ref{ablation}.

\vspace{-1em}

\subsubsection{Detector Free Method}
Given two images $I^A$ and $I^B$, our method accepts a detector-free backbone $F_{df}$ that processes the entire image to output semi-dense correspondences without relying on explicit keypoint detection. Instead, dense feature extraction is performed using deep neural networks. The output features $f_c^{A}$ and $f_c^{B}$ are then matched using similarity metrics, which are generated to quantify the confidence and select the final correspondences. In our implementation, we use AspanFormer~\cite{chen2022aspanformer} as our detector-free backbone.

\vspace{-1em}

\subsubsection{Sampson Distance}
Given a fundamental matrix $F_m$ and the correspondence $P^A_i, P^B_j$, the Sampson Distance $d(i,j)$ is the distance of $P^A_i, P^B_j$ to their true epipolar line, which is given by:
\vspace{-0.5 em}
\begin{equation}
\footnotesize
    d(i,j) = \frac{\left( \mathbf{P}_{j}^{B}{}^\top \mathbf{F_m} \mathbf{P}_{i}^A \right)^2}{(\mathbf{F_m}\mathbf{P}_{i}^A)_1^2 + (\mathbf{F_m}\mathbf{P}_{i}^A)_2^2 + (\mathbf{F_m}^\top\mathbf{P}_{j}^B)_1^2 + (\mathbf{F_m}^\top\mathbf{P}_{j}^B)_2^2}
    \label{eq:sampson},
\end{equation}

where point \( \mathbf{P}_i^A = (x_i, y_i, 1)^\top \) and point \( \mathbf{P}_i^B = (x_j, y_j, 1)^\top \) are two points in $I^A$ and $I^B$ respectively and \( (\mathbf{F_m}\mathbf{P}_{i}^A)_k^2 \) is the k-th entry of the vector \( (\mathbf{F_m}\mathbf{P}_{i}^A) \). Therefore, the Sampson distance~\cite{Hartley:2003:MVG:861369} gives the 
first-order approximation of the reprojection error. The error is minimized when two points are a perfect match.

\subsection{Geometry Aware Optimization Module}
\label{feature matching}

Our proposed optimization module aims to enforce geometric consistency among correspondences and guide the feature matching with geometric information. We utilize $M_s$ from $F_{db}$ to initialize the fundamental matrix. With dense features $f_c^{A}$ and $f_c^{B}$ from $F_{df}$, our module first computes a confidence map $P_c$ using the similarity metrics.
Then the Sampson Distance~\cite{Hartley:2003:MVG:861369} is used as a geometry constraint of an image pair to iteratively select and reassign the correspondences by updating $P_c$. Such constraints can reduce mismatches and address missed matches due to insufficient appearance features and repeating texture patterns.

\textbf{Initial Step}
Given the features $f_c^{A}$ and $f_c^{B}$, the initial confidence matrix is obtained with a dual-softmax operator. For the \( i \)-th feature in $I^A$ and the \( j \)-th feature in $I^B$, the similarity score is calculated as
\begin{equation}
 S(i, j) = \langle \mathbf{f}_c^{A}(i), \mathbf{f}_{c}^{B}(j) \rangle ,
\end{equation}
where \( \langle a, b \rangle \) 
denotes the dot product. Then the confidence level \( P_c \) is calculated as
\begin{equation}
P_c(i, j) = \text{softmax}(S(i, \cdot))_j \cdot \text{softmax}(S(\cdot, j))_i ,
\end{equation}
Based on the initial confidence matrix, 
\vspace{-1 em}

\begin{equation}
    \small
    M_c = \{(\tilde{i},\tilde{j})|\forall(\tilde{i},\tilde{j})\in MNN(P_c), P_c(\tilde{i},\tilde{j}) \geq \theta_{iter}\} .
    \label{eq:MNN}
\end{equation}

we follow LoFTR~\cite{sun2021loftr} 
to apply mutual nearest neighbor ($MNN$) criteria using equation~\ref{eq:MNN} to obtain the initial matches. The row index $i$ represents the index of the patch from $I^A$ and the column index $j$ represents the index from $I^B$. $(\tilde{i},\tilde{j})$ is a match if and only if the confidence level is the maximum of both $i$th row and $j$th column. We choose points where their confidences are larger than $\theta_{iter}$ as matches to enter iterative steps.

The initial fundamental matrix $F_{m}$ is calculated using the normalized 8-points method if there exist enough good anchor points in $M_s$, which is at least 10 points with confidence larger than 0.5. Otherwise, the initial fundamental matrix will be calculated using the top half matches from equation ~\ref{eq:MNN}.

\textbf{Iterative Steps}
Our iterative steps begin with initialized coarse matches $M_s$ and fundamental matrix $F_m$ and update them in each iteration. The Sampson distance error is calculated for all initial matches according to Equation~\ref{eq:sampson}.
Geometric confidence is then estimated from the Sampson distances as 
\begin{equation}
    P_d(i, j) = \text{sigmoid}(\text{relu}(\tau-d(i, j)),
\end{equation}
where \( \tau \) is a distance threshold, and smaller values of \( \tau \) will lead to more strict constraints. The geometric confidence for correspondences that have distances larger than the distance threshold will be 0.5. In other words, the confidence of matches that violate the geometry constraint will be penalized.

Finally, the confidence map is updated by the geometric confidence map by a given weight \( w \). Therefore, the final confidence is
\begin{equation}
    P_c(i, j)_i =  P_c(i, j)_{i-1} \cdot P_d(i, j)_{i-1} \cdot w.
\end{equation}
At every iteration, the updated confidence map is normalized using min-max normalization to avoid extreme values. When the iteration ends, the module will select the final coarse-level matches 
$M_{c_{final}}$ based on the final confidence map from the last iteration with mutual nearest neighbor criteria according to equation \ref{eq:MNN}, but with a different threshold $\theta_{c}$ to avoid selecting a large amount of correspondences with very small confidences.  
%\haiwei{why a different threshold? feel like we missed some explanation here}. 
The refined matches $M_{c_{final}}$ are further fed into a correlation-based refinement block, which is the same with LoFTR~\cite{sun2021loftr}, to obtain the final matching results.

\begin{figure*}[!ht]
    \centering
    \includegraphics[width=0.97\linewidth]{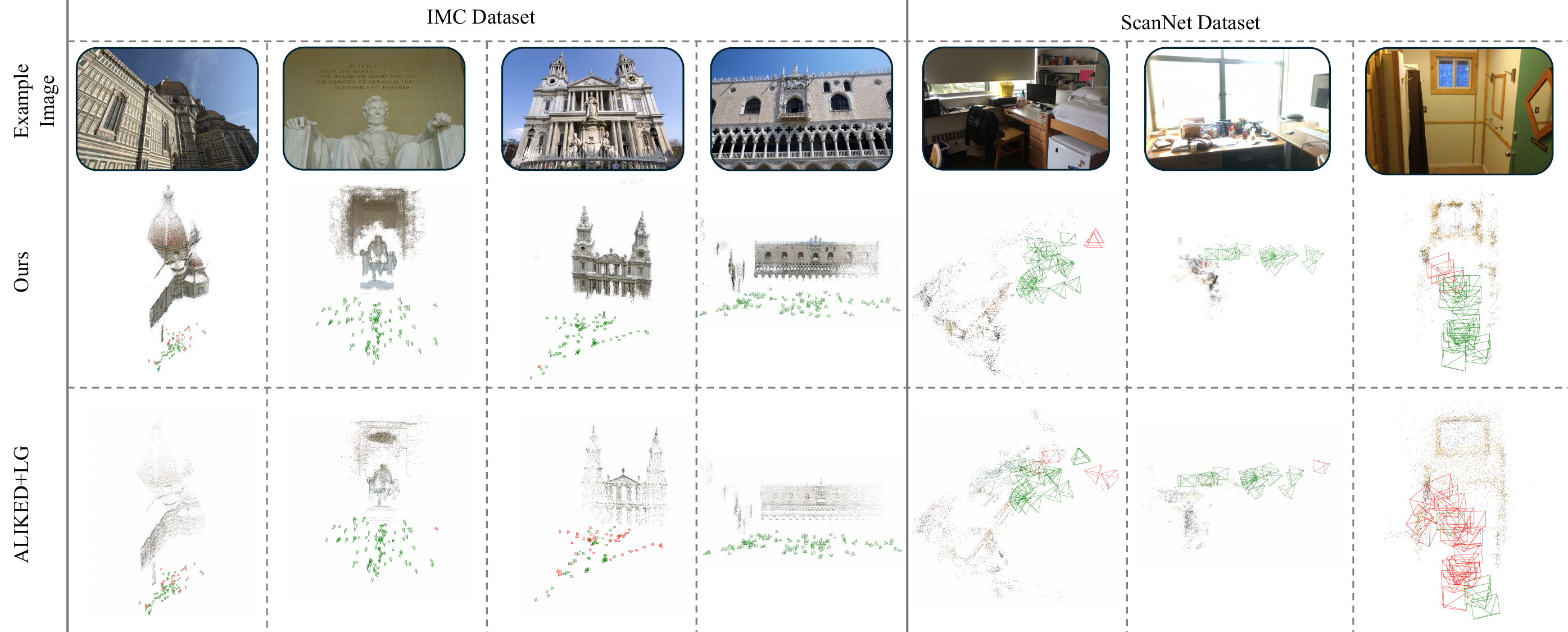}
    \caption{\textbf{Qualitative Results}. Our method is qualitatively compared with ALIKED~\cite{zhao2023aliked} + LG~\cite{lindenberger2023lightglue} on multiple scenes. Green cameras have less than 3\degree absolute pose error, while red cameras have an error larger than 3\degree. More results can be found in supplementary material.}
    \label{fig:imc_rec}
\end{figure*}

\subsection{Structure from Motion Reconstruction}
The final matching results are fed into COLMAP~\cite{schoenberger2016sfm} for estimating structure from motion. By integrating our robust and dense matches into COLMAP~\cite{schoenberger2016sfm}, our method is shown to enhance the accuracy and completeness of the camera poses and reconstructed point cloud, resulting in superior 3D models, even in challenging scenarios with significant viewpoint changes, as discussed in Section~\ref{sub:3d}. 

\subsection{Implementation Details}
The Sampson Distance threshold \( \tau \) is set to 10 which allows a few pixels of error. The iteration of the optimization module is set to 10 and the weight \( w \) is 1.2. The confidence threshold \(\theta_{iter}\) is set to a small value of 0.01. The idea is to conduct a comprehensive search across the space. The effect of different hyperparameters will be discussed in Section~\ref{ablation}.
\vspace{-1em}
\section{Experiments}

In this section, we demonstrate the performance of our method in the SfM reconstruction task in terms of the accuracy of camera pose estimation and its application in challenging real-world scenarios. We also conduct a comprehensive ablation study. All experiments were run on an RTX 4090 GPU with 24GB VRAM and a 13th Gen Intel® Core™ i9-13900KF CPU with 32 cores and 126GB RAM.

\subsection{Dataset} 
We utilize the Image Matching Challenge (IMC) 2021~\cite{Jin2020} and ScanNet~\cite{dai2017scannet} benchmark datasets to evaluate our method's performance in SfM reconstruction for outdoor and indoor settings respectively. As there are no publicly available air-to-ground datasets, we collected two additional air-to-ground datasets to demonstrate our method's ability to find reliable correspondences in this challenging scenario.

The IMC Phototourism benchmark~\cite{Jin2020} encompasses multiple large-scale outdoor scenes. The test set includes 9 sites, and each set has 100 images. The ground truth or pseudo ground truth camera poses are obtained from COLMAP~\cite{schoenberger2016sfm} on the entire collection of images and then sampled into test sets. ScanNet~\cite{dai2017scannet} contains 1613 monocular indoor sequences with ground truth poses and depth maps. We use the data split of the test set following SuperGlue~\cite{sarlin20superglue} and LoFTR~\cite{sun2021loftr}, which includes 15 sequences. The primary challenge lies in accurately recovering the camera poses from in-the-wild images with significant viewpoint and illumination changes. For all test scenes, the images are considered unordered.

The first collected air-to-ground dataset consists of 16 UAV images and 21 ground images, while the second dataset contains 16 UAV images and 24 ground images. The primary challenge for these datasets is to accurately register UAV images with ground view images. For all scenes, the images are considered unordered and are matched exhaustively. When aerial images are matched with ground images, to avoid failure in finding correspondences due to resizing, the aerial images are cropped into quarters plus a center region to generate five pairs. The matches are subsequently merged to obtain the final matches. This strategy is used for all methods.

\begin{table}[!ht]
    \centering
    \begin{tabular}{l>{\centering\arraybackslash}m{1cm}>{\centering\arraybackslash}m{1cm}>{\centering\arraybackslash}m{1cm}}
    \Xhline{1pt}
    \noalign{\smallskip}
    \multirow{2}{*}{\textbf{Method}}  & \multicolumn{3}{c}{\textbf{Pose Estimation AUC} $\uparrow$} \\
    \noalign{\smallskip}
    \cline{2-4}
    \noalign{\smallskip}
    & @$3\degree$ & @$5\degree$ & @$10\degree$ \\
    \noalign{\smallskip}
    \Xhline{1pt}
    \noalign{\smallskip}
    \textit{SIFT~\cite{Lowe:2004:DIF:993451.996342}+NN}  & 51.5 & 65.3 & 78.9 \\
    \textit{SP~\cite{superpoint2018}+SG~\cite{sarlin20superglue}} & 59.6 & 71.0 & 83.3 \\
    \textit{ALIKED~\cite{zhao2023aliked}+LG~\cite{lindenberger2023lightglue}} & \underline{66.0} & \underline{76.8} & \underline{87.6} \\
    \textit{DKM~\cite{edstedt2023dkm}} & 42.3 & 58.5 & 76.3 \\
    \textit{RoMa~\cite{edstedt2024roma}} & 36.3 & 51.6 & 69.5 \\
    \textit{Ours} & \textbf{69.1} & \textbf{80.5} & \textbf{90.1} \\
    \Xhline{1pt}
    \end{tabular}
    \caption{Estimated pose errors on IMC 2021 phototourism~\cite{Jin2020} in outdoor scences. Results are averaged across all scenes.}
    \label{imc_pose}
    \vspace{-1em}
\end{table}

\subsection{Metrics and Comparing Methods} 
We report the metric area-under-curve (AUC) to evaluate camera pose accuracy for IMC 2021~\cite{Jin2020} and ScanNet~\cite{dai2017scannet}. Specifically, for every reconstructed model, we aligned the predicted camera poses with the ground truth camera poses. We then computed the absolute angular translation and rotation errors. The AUC selects the maximum between these two errors and outputs the area under the accuracy threshold curve. As the collected air-to-ground datasets lack ground-truth camera poses, we follow~\cite{schoenberger2016sfm} to report the number of registered images, the number of 3D points, the average track length, and the average projection error (px) in this setting to evaluate the methods by the completeness of the reconstructed model.

\begin{table}[!ht]
    \centering
    \begin{tabular}{l>{\centering\arraybackslash}m{1cm}>{\centering\arraybackslash}m{1cm}>{\centering\arraybackslash}m{1cm}}
    \Xhline{1pt}
    \noalign{\smallskip}
    \multirow{2}{*}{\textbf{Method}}  & \multicolumn{3}{c}{\textbf{Pose Estimation AUC} $\uparrow$}\\
    \noalign{\smallskip}
    \cline{2-4}
    \noalign{\smallskip}
    & @$3\degree$ & @$5\degree$ & @$10\degree$ \\
    \noalign{\smallskip}
    \Xhline{1pt}
    \noalign{\smallskip}
    \textit{SIFT~\cite{Lowe:2004:DIF:993451.996342}+NN}  & 18.6 & 24.1 & 30.4 \\
    \textit{SP~\cite{superpoint2018}+SG~\cite{sarlin20superglue}} & \underline{45.9} & \underline{61.1} & 73.8 \\
    \textit{ALIKED~\cite{zhao2023aliked}+LG~\cite{lindenberger2023lightglue}} & 44.0 & 59.5 & \underline{74.3} \\
    \textit{DKM~\cite{edstedt2023dkm}} & 23.5 & 39.0 & 60.7 \\
    \textit{RoMa~\cite{edstedt2024roma}} & 28.5 & 47.3 & 67.6 \\
    \textit{Ours} & \textbf{46.0} & \textbf{63.4} & \textbf{77.6} \\
    \Xhline{1pt}
    \end{tabular}
    \caption{Estimated pose errors on ScanNet~\cite{dai2017scannet} in indoor scences. Results are averaged across all scenes.}
    \label{mega_pose}
    \vspace{-1em}
\end{table}

We compared our method against four different methods:
SIFT~\cite{Lowe:2004:DIF:993451.996342}+NN, which is the default feature matching method in COLMAP~\cite{schoenberger2016sfm, schoenberger2016mvs}; SuperPoint~\cite{superpoint2018}+Superglue~\cite{sarlin20superglue}, a robust method widely used in SfM; ALIKED~\cite{zhao2023aliked}+LightGlue~\cite{lindenberger2023lightglue}, a state-of-the-art detector-based feature matching method; DKM~\cite{edstedt2023dkm} and RoMa~\cite{edstedt2024roma}, two state-of-the-art dense feature matching methods. We omit RoMa~\cite{edstedt2024roma} for air-to-ground datasets since DKM~\cite{edstedt2023dkm} appears to be superior to RoMa~\cite{edstedt2024roma} in terms of 3D reconstruction for outdoor scenes.

In order to reduce the computational time, the number of matches per pair is limited to 4K for all methods. Bold text indicates the best results, and underlined text indicates the second best.

\subsection{Comparison to State of the Art Methods}
\label{sub:3d}

\begin{figure*}[!ht]
    \centering
    \includegraphics[width=0.97\linewidth]{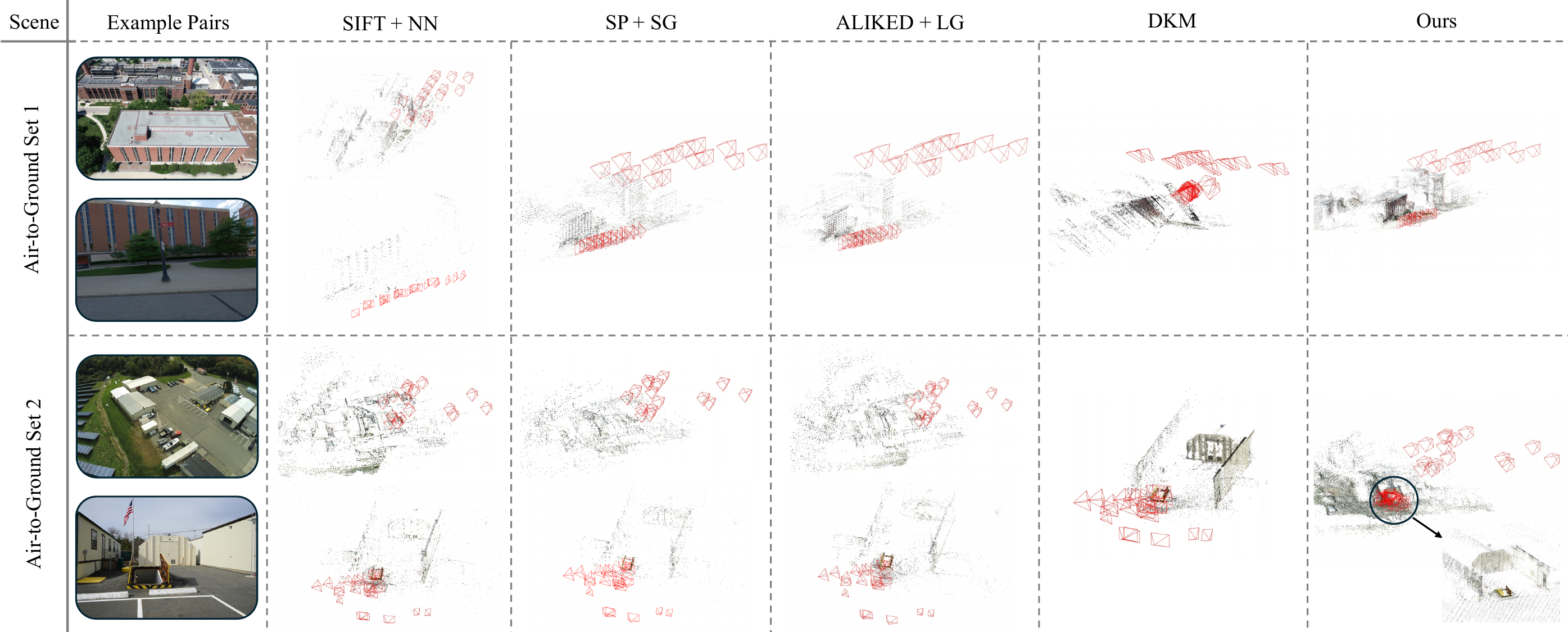}
    \caption{\textbf{Qualitative Results}. Our method is qualitatively compared with other feature matching methods on collected air-to-ground datasets. Red cameras are recovered poses.}
    \label{fig:air_ground}
\end{figure*}

\begin{table*}
    \centering
    \resizebox{1\textwidth}{!}{
    \setlength{\tabcolsep}{4pt}
    \begin{tabular}{lcccccccccccc}
        \toprule
        \multirow{2}{1.5cm}[-.4em]{\textbf{Metrics}}
        & \multicolumn{5}{c}{\textbf{Air to Ground Set 1 (37 Images)}} & \multicolumn{5}{c}{\textbf{Air to Ground Set 2 (40 Images)}} \\
        \cmidrule(lr){2-6} \cmidrule(lr){7-11}
        & \textit{SIFT+NN} & \textit{SP+SG}  & \textit{ALIKED+LG} & \textit{DKM} & \textit{Ours} & \textit{SIFT+NN} & \textit{SP+SG}  & \textit{ALIKED+LG} & \textit{DKM} & \textit{Ours} \\
        
        \midrule
        
        \multirow{1}{*}{\# Registered $\uparrow$}
        & 16/18 & 36 & 36 & 36 & \textbf{37} & 16/24 & 16/24 & 16/24 & 25 & \textbf{40} \\
        
        \multirow{1}{*}{\# 3D Points $\uparrow$}
        & 11.2K/1.27K & 11.5K & 12.5K & \underline{22.1K} & \textbf{53.2K} & 13.0K/11.3K & 5.52K/3.60K & 7.95K/8.75 & \underline{48.0K} & \textbf{74.4K} \\

        \multirow{1}{*}{Avg. Track Length $\uparrow$}
        & 3.4/2.7 & 3.08 & \textbf{3.7} & 2.2 & \underline{3.1} & 3.5/3.6 & 3.4/3.2 & 3.9/3.5 & 2.8 & \textbf{3.6} \\

        \multirow{1}{*}{Avg. Reproj. Error $\downarrow$}
        & 1.05/0.99 & 1.32 & 1.25 & \textbf{0.81} & \underline{1.22} & 1.15/1.00 & 1.39/1.36 & 1.35/1.33 & \textbf{0.84} & \underline{1.28} \\
        \bottomrule
    \end{tabular}
}
\caption{ 
        SfM results with different feature matchers on Air-to-Ground datasets. For methods that generate 2 separate models, we report them as ``air model/ground model".
    }
    \vspace{-0.5em}
    \label{air_ground_rec}
\end{table*}

SfM results using our feature matching method on both the IMC 2021~\cite{Jin2020} and the ScanNet~\cite{dai2017scannet} datasets have shown significant improvements in performance. As shown in Table~\ref{imc_pose} and Table~\ref{mega_pose}, our feature matcher contributes to the highest average pose estimation AUC in both datasets. Specifically, Table~\ref{imc_pose} indicates an increase of +3.1 in AUC@$3\degree$ for estimated camera poses compared to ALIKED~\cite{zhao2023aliked} + LG~\cite{lindenberger2023lightglue}, as can be visually shown in Figure~\ref{fig:imc_rec}. Similar gains are observed for ScanNet~\cite{dai2017scannet}, as shown in Table~\ref{mega_pose}. These findings underscore the efficacy of our method in achieving superior SfM results. 

As shown in Table~\ref{air_ground_rec} and Figure~\ref{fig:air_ground}, compared to other methods, our method is able to register all images and is the only method that can align the ground images with UAV images for set 2. We have noticed that reconstructed camera poses using DKM~\cite{edstedt2023dkm} in set 1 are apparently wrong with ground images pointing down and speculate that it is due to the lack of multi-image matches for feature tracking. In addition, DKM~\cite{edstedt2023dkm} also fails in reconstructing a model for UAV images in set 2. These results further validate the practical utility and effectiveness of our feature matcher in real-world scenarios.

\subsection{Ablation Study}
\label{ablation}

We have conducted an ablation study to investigate the impact of different parameters and design choices on both feature matching and 3D reconstruction. 

\textbf{Ablation study on Homography Estimation}. Feature matching is evaluated on the widely adopted HPatches dataset~\cite{hpatches_2017_cvpr}. HPatches~\cite{hpatches_2017_cvpr} contains a total of 108 sequences with significant illumination changes and large viewpoint changes. We follow the evaluation protocol of LoFTR~\cite{sun2021loftr} to resize the shorter size of the image to 480 and report AUCs at 3 different thresholds. Table~\ref{tab:ablation} shows results on different parameters and design choices on HPathes~\cite{hpatches_2017_cvpr}. The ``Baseline'' denotes only using ASpanFormer~\cite{chen2022aspanformer}. The results show the effects of different parameters in feature matching. Additionally, the full method outperforms other design variations, confirming the robustness and effectiveness of our approach in enhancing feature matching accuracy. More experiments related to feature matching can be found in supplementary material.
 
\begin{table}[!ht]
    \centering
    \resizebox{0.47\textwidth}{!}{%
    \begin{tabular}{ccccc}
    \toprule
    Types & Values & \multicolumn{3}{c}{AUC} \\ 
    \cmidrule(lr){3-5}
    & & @$3 px$ & @$5 px$ & @$10 px$ \\
    \midrule
    \multirow{3}{*}{(1) Updating weight} & $w=1.2$ & \textbf{70.3} & \textbf{79.5} & \textbf{88.0} \\
    & $w=1.5$ &  70.2 & 79.4 & \textbf{88.0} \\ 
    & $w=1.8$ &  \textbf{70.3} & \textbf{79.5} & \textbf{88.0} \\
    \midrule
    
    \multirow{4}{*}{(2) Number of iterations} & No optimization. & 68.9 & 78.0 & 87.4 \\
    & 5 iter & 68.2 & 78.1 & 87.1  \\
    & 10 iter & \textbf{70.3} & \textbf{79.5} & \textbf{88.0} \\
    & 20 iter & 68.2 & 78.1 & 87.1 \\
    \midrule
    \multirow{4}{*}{\begin{tabular}[c]{@{}r@{}}(3) Distance threshold\end{tabular}} & \( \tau \) = 1 & 69.0 & 78.5 & 87.0 \\
    & \( \tau \) = 10 & 70.3 & 79.5 & \textbf{88.0} \\
    & \( \tau \) = 50 & \textbf{71.2} & \textbf{79.8} & 87.6    \\
    & \( \tau \) = 200 &  70.7 & 79.73 & 87.7 \\
    \midrule
    \multirow{4}{*}{(4) Designs}& Full method & \textbf{70.3} & \textbf{79.5} & \textbf{88.0} \\
    & No anchor points & 67.4 & 77.6 & 86.8 \\
    & No optimization & 68.9 & 78.0 & 87.4 \\
    & Baseline & 66.1 & 75.9 & 84.8  \\
    \bottomrule
    \end{tabular}%
    }
    \caption{Ablation study of our proposed method in Homography Estimation on Hpatches~\cite{hpatches_2017_cvpr}.}
    \label{tab:ablation}
\end{table}

\begin{table}[!ht]
    \centering
    \resizebox{0.37\textwidth}{!}{%
    \begin{tabular}{l>{\centering\arraybackslash}l>{\centering\arraybackslash}m{2cm}>{\centering\arraybackslash}m{2cm}>{\centering\arraybackslash}m{2cm}}
    \Xhline{1pt}
    \noalign{\smallskip}
    \multirow{2}{*}{\textbf{SG}} & \multirow{2}{*}{\textbf{Geo}}  & \multicolumn{3}{c}{\textbf{Pose Estimation AUC}$\uparrow$} \\
    \noalign{\smallskip}
    \cline{3-5}
    \noalign{\smallskip}
    & & @$3\degree$ & @$5\degree$ & @$10\degree$ \\
    \noalign{\smallskip}
    \Xhline{1pt}
    \noalign{\smallskip}
    &  & 50.0 & 62.4 & 75.6 \\
    \checkmark &  & 56.0 & 66.3 & 76.0 \\
    & \checkmark & 57.0 & 69.8 & 82.1\\
    \checkmark & \checkmark & \textbf{58.6} & \textbf{70.8} & \textbf{82.9} \\
    \Xhline{1pt}
    \end{tabular}
    }
    \caption{Ablation study of our proposed method in 3D reconstruction of sampled images from IMC 2021 phototourism dataset. Results are averaged across all scenes.}
    \label{ablation_rec}
\end{table}

\begin{table}[!ht]
    \centering
    \resizebox{0.45\textwidth}{!}{%
    \begin{tabular}{l>{\centering\arraybackslash}l>{\centering\arraybackslash}m{2cm}>{\centering\arraybackslash}m{2cm}>{\centering\arraybackslash}m{2cm}>{\centering\arraybackslash}m{2cm}}
    \Xhline{1pt}
    \noalign{\smallskip}
    \textbf{SG} & \textbf{Geo}  & \textbf{\# Registered} & \textbf{\# Points $\uparrow$} & \textbf{{Reproj. Err(px) $\downarrow$}} & \textbf{{Mean Track Length $\uparrow$}}  \\
    \noalign{\smallskip}
    \Xhline{1pt}
    \noalign{\smallskip}
    &  & 15 & 12.7K & \textbf{0.334} & 3.3\\
    \checkmark &  & \textbf{20} & \textbf{72.9K} & 0.581 & 5.3 \\
    & \checkmark & 19 & 55.9K & 0.468 & 5.0 \\
    \checkmark & \checkmark & \textbf{20} & 64.2K & 0.508 & \textbf{5.6} \\
    \Xhline{1pt}
    \end{tabular}
    }
    \caption{Ablation study of our proposed method in 3D reconstruction of sampled images from IMC 2021 phototourism dataset. Results are averaged across all scenes.}
    \label{ablation_auc}
\end{table}

\begin{table}[!ht]
    \centering
    \resizebox{0.37\textwidth}{!}{%
    \begin{tabular}{l>{\centering\arraybackslash}l>{\centering\arraybackslash}m{1.2cm}>{\centering\arraybackslash}m{1.2cm}>{\centering\arraybackslash}m{1.2cm}}
    \Xhline{1pt}
    \noalign{\smallskip}
    \multirow{2}{*}{\textbf{Detector Based Method}} & \multicolumn{3}{c}{\textbf{Pose Estimation AUC}$\uparrow$} \\
    \noalign{\smallskip}
    \cline{2-4}
    \noalign{\smallskip}
    & @$3\degree$ & @$5\degree$ & @$10\degree$ \\
    \noalign{\smallskip}
    \Xhline{1pt}
    \noalign{\smallskip}
     SP~\cite{superpoint2018}+SG~\cite{sarlin20superglue} & \textbf{69.1} & \textbf{80.5} & \textbf{90.1} \\
     ALIKED~\cite{zhao2023aliked}+LG~\cite{lindenberger2023lightglue} & 65.3 & 78.3 & 89.1 \\
    \Xhline{1pt}
    \end{tabular}
    }
    \caption{Ablation study of using different detector-based backbone in SfM on IMC dataset~\cite{Jin2020}.}
    \vspace{-1em}
    \label{ablation_detector}
\end{table}

\textbf{Ablation study on 3D reconstruction}. In order to comprehensively assess the efficacy of our proposed components within the SfM system, we also measure the impact of the aforementioned design changes on 3D reconstruction. To this end, we randomly sample 20 images from each scene of IMC 2021 phototourism datasets~\cite{Jin2020} and compare the SfM reconstruction results based on pose estimation. We follow IMC~\cite{Jin2020} to report relative pose errors for all possible pairs since we are using subsets for this experiment. All results are averaged across all scenes and all methods are evaluated on the same subset of data. 

``SG'' denotes anchor points from SuperGlue~\cite{sarlin20superglue}; if the optimization (``Geo'') module is not enabled, anchor points are naively concatenating to final matches. If there are no anchor points, the optimization (``Geo") module is initialized with the top half most confident matches from only the detector-free backbone. As shown in Tables~\ref{ablation_rec} and \ref{ablation_auc}, the results again demonstrate the effectiveness of our proposed method as it brings improvement over the baseline methods in the accuracy of SfM reconstruction.

\textbf{Ablation study on detector-based backbones}. Following the setting of Section~\ref{sub:3d}, Table~\ref{ablation_detector} compares the performance of using two detector-based methods on the IMC dataset~\cite{Jin2020}. This demonstrates that SP\cite{superpoint2018} + SG\cite{sarlin20superglue} is more effective in providing priors in guiding dense matching.

\section{Limitation and Future Work}

\begin{table}[!ht]
    \centering
    \resizebox{0.45\textwidth}{!}{%
    \begin{tabular}{ccccc}
    \Xhline{1pt}
     & Image Retrieval & Feature Matching & Reconstruction \\
    \midrule
    Avg. Runtime (s) & 3 & 1769 & 1478 \\
    \bottomrule
    \end{tabular}
    }
    \caption{Average runtime analysis on IMC 2021~\cite{Jin2020}. Each site contains around 100 images and image retrieval proposes an average of 1750 pairs.}
    \label{runtime}
\end{table}
\vspace{-1 pt}
The main limitation of our method is efficiency, due to the computational cost of the transformer-based architecture and the significant number of correspondences produced, as shown in Table~\ref{runtime}. Utilizing a newly developed backbone model with a focus on efficiency, such as Efficient LoFTR~\cite{wang2024eloftr}, can improve it with slightly degraded performance.

Due to our method's reliance on the backbone model to provide initial matches for fundamental matrix estimation and its heuristic design, our method could fail with poor initial matches from backbone models. Nevertheless, our evaluations against baselines demonstrate a marked improvement in accuracy, as well as the capability to find reliable matches in challenging scenarios. Furthermore, we believe the accuracy of SfM reconstruction can be further improved by adopting multi-view refinement techniques, such as those proposed by pixSfM~\cite{lindenberger2021pixsfm}.

\section{Conclusion}
We propose the geometry-aware optimization module, an innovative optimization approach that bridges detector-based matcher and detector-free matcher to improve SfM reconstruction and accuracy of estimated camera poses. This module leverages epipolar constraints, iteratively optimizes for the best relative pose, and applies Sampson distance weighting to refine match precision. The introduction of geometry optimization enhances the system's ability to handle challenging scenarios, including drastic changes in perspective, scale, and illumination. Our method is shown to be also effective in 3D reconstruction for large baselines, a scenario where conventional methods often fail and perform unsatisfied. 
\section{Acknowledgement}
Supported by the Intelligence Advanced Research Projects Activity (IARPA) via Department of Interior/ Interior Business Center (DOI/IBC) contract number 140D0423C0075. The U.S. Government is authorized to reproduce and distribute reprints for Governmental purposes notwithstanding any copyright annotation thereon. Disclaimer: The views and conclusions contained herein are those of the authors and should not be interpreted as necessarily
representing the official policies or endorsements, either expressed or implied, of IARPA, DOI/IBC, or the U.S. Government.
{
    \small
    \bibliographystyle{ieeenat_fullname}
    \bibliography{main}
}
\end{document}